# Languages for Smart and Computable Contracts

*Christopher D. Clack*
*Centre for Blockchain Technologies*
*Department of Computer Science, UCL*
clack@cs.ucl.ac.uk
8th April 2021

*Abstract*

Smart Contracts use computer technology to automate the performance of aspects of commercial agreements. Yet how can there be confidence that the computer code is faithful to the intentions of the parties? To understand the depth and subtlety of this question requires an exploration of natural and computer languages, of the semantics of expressions in those languages, and of the gap that exists between the disciplines of law and computer science. Here we provide a perspective on some of the key issues, explore some current research directions, and explain the importance of language design in the development of reliable Smart Contracts, including the specific methodology of Computable Contracts.

Keywords: Smart Legal Contracts, Computable Contracts, Domain Specific Languages, Controlled Natural Language

*Introduction*

The field of Smart Contracts is broad, with multiple and sometimes contradictory definitions of what is considered to be a Smart Contract. Various factors have contributed to this complexity, including a technical divergence that occurred when the term was used to describe stored procedures in the Ethereum blockchain,[1] which conflicted with the original definition (pre-dating Ethereum by 17 years) that aimed to automate commercial agreements in general, regardless of technology platform.[2] Significant research has been, and continues to be, conducted on general Smart Contracts that are not necessarily linked to blockchains. Another factor is the broad range of disciplines involved, including computer science, law, logic and linguistics.

Moreover, there is a fundamental conflict inherent in the term 'Smart Contract', which brings together the two disparate and highly specialised disciplines of computer science ('Smart', as in 'smart phone') and law ('Contracts'). Although there may be a surface similarity between a written contract and a computer program (both of which are carefully structured and may contain *inter alia* definitions, descriptions of actions to be taken, and conditional logic), there are substantial differences between the language, culture and perspective of lawyers and computer scientists.

---

[1] *Buterin, V. (2013) A Next Generation Smart Contract & Decentralized Application Platform. Whitepaper. Ethereum Foundation.*
[2] *Szabo, N. (1996). Smart contracts: building blocks for digital markets. EXTROPY: The Journal of Transhumanist Thought,(16), 18, 2.*

It is to be expected that differences of opinion and commercial or political imperatives will contribute to continuing debate about definitions for a while longer. Yet there is also a growing acknowledgement that the potential for misunderstanding impedes progress, and more general definitions of the term have been created in order to obtain broader consensus. An example is the widely-cited portmanteau definition from Clack et al:[3]

> "A smart contract is an agreement whose execution is both automatable and enforceable. Automatable by computer, although some parts may require human input and control. Enforceable by either legal enforcement of rights and obligations or tamper-proof execution."

Even this definition has its problems: it uses the term 'execution' in a computing sense (the running of computer code), whereas from a lawyer's perspective perhaps 'performance' might have been a better choice. Furthermore, definitions of the terms 'automatable' and 'enforceable' are not given, leaving room for differing interpretations; though this may also explain the popularity of this definition.

The automation of commercial agreements requires that some[4] or all of the intentions of the parties should be expressed in computer code, yet how can there be confidence that the computer code is faithful to those intentions? It is not sufficient to check whether the code is correct solely with respect to those aspects being automated, but also to check that there is no deviation from any aspect of the entire agreement.[5] The matter is more acute where it is intended to automate the performance of some or all of a legal agreement (a 'smart legal contract' as defined by Stark[6], drawing a contrast with the automation of agreements that are not contracts), especially one of high value and potentially very long term (perhaps stretching into decades, as for example with some financial contracts) and whose legal documentation may have substantial size and complexity.

The following sections explore aspects of this question ("is the code faithful to the agreement?") in the context of smart legal contracts, to demonstrate some of the depth and subtlety of the issues at play. The aim is to provide a better understanding of the different ways in which technology may be used to convert aspects of a legal agreement into computer code[7] and of the issues that arise when attempting to validate the behaviour of the code. Much of the discussion will focus on language:

- The specialised natural languages used by computer scientists and lawyers;
- A variety of synthetic languages used by computer scientists.

---

[3] *Clack, C. D., Bakshi, V. A., & Braine, L. (2016). Smart contract templates: foundations, design landscape and research directions. arXiv preprint arXiv:1608.00771.*

[4] *The selection of which aspects of an agreement to automate is beyond the scope of this article, and has been covered in a specific context elsewhere: see op. cit. Footnote 34.*

[5] *For example, code that automates the calculation and performance of payments and deliveries might fail to implement an appropriate grace period in the case of delayed payment.*

[6] *Stark, J. (2016). Making sense of blockchain smart contracts. Coindesk.com. http://www.coindesk.com/making-sense-smart-contracts/*

[7] *This is part of what Allen calls the 'technology stack' (see Footnote 8).*

Section 1 introduces various terms of art from computer science and the 'language stack' of different languages that may be used during the conversion from the agreement to the instructions that control a computer. Section 2 discusses aspects of natural and formal expression; this includes issues that arise when converting from agreement to code, such as whether the code defines or implements contractual obligations, and strategies for validating repeated updates to the code or the agreement. Section 3 analyses issues that arise with respect to the semantics of the agreement and the semantics of the code, and Section 4 discusses research developments and directions in smart contract methodology using 'Computable Contracts' – including markup languages, templates, domain specific languages and controlled natural languages.

## 1    The Language Stack

The automation by computer of selected aspects of a legal agreement requires code to be written, and that code must be verified and validated.  Here we use the term 'verified' to mean an internal process whereby computer code is checked for technical correctness (is the programmer building the code right?), whereas the term 'validation' means an external process whereby previously verified code is checked for whether it is faithful to the agreement (is the programmer building the right code?).

Validation can be substantially more difficult than verification, and establishing whether the code is faithful to the agreement may require (for example) that all possible behaviours of the code (i.e. sequences of actions and calculations, with timings, performed by the code whether or not in response to input data) correctly automate exactly what needs to be automated (no more and no less) and without conflicting with any aspect of the overall agreement.  For complex code it may be impossible to engage in exhaustive testing of all behaviours for all inputs and outputs because there are too many combinations, and so formal analysis based on program semantics may be used (see Section 3). The person or team that is responsible for validation therefore needs to have a full understanding of both the agreement and computer science.  The former may for example require a full understanding of the relevant law and of standard business practices for the relevant sector.  The latter requires *inter alia* a full understanding of the 'language stack' as explained below.

This article focuses on written contracts. Whilst the legal documentation is likely to be a key resource, and a large part of the effort of creating and validating code will focus on conversion from the legal documentation to the code, it is not the only source of information about the agreement. Allen[8] argues that a contract can usefully be viewed as a complex legal institutional entity comprising a 'stack' of interacting legal and technological 'layers'.  He introduces the term 'contract stack' to describe

---

[8] *Allen, J. G. (2018). Wrapped and stacked:'smart contracts' and the interaction of natural and formal language. European Review of Contract Law, 14(4), 307-343.*

this structure and gives the example of a written contract whose 'contract stack' comprises:[9]

> "(i) the spoken words through which the contractual terms were negotiated and against which the text was drafted, (ii) the written text, and (iii) legal rules implying terms and governing construction"

Allen relates this notion of 'contract stack' to the 'technology stack' of the underlying elements (such as the languages and software products) of a computer application. The appearance of languages in the technology stack is extremely important; multiple languages are used in a highly structured manner, and this article therefore introduces the term 'language stack'.

It is essential to understand the 'language stack' in order to understand the many issues that arise when validating whether the automation of an agreement is faithful to that agreement. A computer can only carry out instructions provided in an **executable language** (see below) comprising sequences of binary digits (0 and 1), and that executable language is very far removed from the human-readable code produced by programmers. The conversion from human-readable code to executable code proceeds via multiple intermediate layers in the language stack, each with its own specialised language(s). Errors and mis-interpretations can occur during the process of conversion down the layers of the language stack to the final bits that control the computer, and it is important to understand why and how these errors and mis-interpretations can occur.

Conversion generally requires a transformation of syntax (words and grammar) while retaining the semantics (throughout this article we use the terms 'meaning', 'semantics' and 'substance' as synonyms), yet the creation of code for a Smart Contract ('smart contract code') will typically require the semantics of only some aspects of the agreement to be converted to code, and may require understanding of aspects of the agreement that are not in the written contract. Furthermore, this process of conversion requires that the semantics of those aspects of an agreement that are to be automated must be known in advance (this does not mean that the semantics must be fully specified for all possible future events, but that the semantics must be known in advance for those future events that are contemplated by the parties – even if the required action in response to an event is "pause and refer to one or both parties for guidance").

When focusing on the legal documentation, the starting point for the language stack is not the code written by a programmer: instead, the starting point is the natural language used in the written documentation. The following brief summary provides sufficient background for subsequent discussion. Relevant terms of art from computer science include 'source code' (an artefact using a human-readable programming or specification language, as defined below) and 'execution' (the performance by a computer of machine-code instructions).

---

[9] *Noting that the extent to which spoken words may be upheld by a court of law may vary according to jurisdiction.*

5- **Natural language.** At the top of the language stack is the natural language (e.g. English) in which aspects of the agreement, including definitions, conditional logic, and actions to be performed, are expressed. Agreements may use sector-specific terms of art and where these terms are complex they may be structured to record the properties of these terms, and the relationships between them, by use of standardised ontologies such as the Financial Industry Business Ontology (FIBO).[10] A **controlled natural language (CNL)**[11] may have even stricter controls on grammar and vocabulary (nouns, verbs, adjectives, adverbs etc.) and may also be a **domain-specific language** (see below).

- **Specification language.** For many decades in computer science and electronic engineering formally-based 'specification languages' have been used for complex descriptions of software and systems.[12] With the increasing computerisation and automation of processes across different sectors, there has been increasing use of specification languages for complex descriptions of commercial systems and processes. Specification languages can also provide formal descriptions ('specifications') of legal agreements. A specification language is not normally a natural language, but rather may be a mathematical or logical formalism,[13,14,15,16,17,18] or may be similar to a programming language and may support computer simulation of contract performance.[19] Conversion between natural and specification languages is typically achieved manually. A specification language could potentially be used before, during or after drafting a natural-language contract and can provide clarity regarding the natural language. For example:

- for analysis to guide the drafting lawyer (such as inconsistency and incompleteness analysis);

- as an intermediate step in translation to a lower layer in the language stack; and/or

- as an agreed definition of the semantics of the natural-language layer, against which the code can be validated (for example, an important role for specification languages is in formal proofs of the correctness of software).[20]

---

[10] https://wiki.edmcouncil.org/

[11] *Kuhn surveys existing English-based CNLs, including 100 CNLs from 1930 to 2014: Tobias Kuhn (2014). A Survey and Classification of Controlled Natural Languages. Association for Computational Linguistics. See also Wyner, A., Angelov, K., Barzdins, G., Damljanovic, D., Davis, B., Fuchs, N., ... & Sowa, J. (2009). On controlled natural languages: Properties and prospects. In International workshop on controlled natural language (pp. 281-289). Springer, Berlin, Heidelberg.*

[12] *For example 'Z notation'. See Abrial, J-R, Schuman, S.A., Meyer, B. (1980), "A Specification Language", in Macnaghten, A. M. and McKeag, R. M. (eds.), On the Construction of Programs, Cambridge University Press, ISBN 0-521-23090-X.*

[13] *Hvitved provides a comparative survey of formal languages and models for contracts, together with his own Contract Specification Language (CSL): Hvitved, T. (2012). Contract formalisation and modular implementation of domain-specific languages (PhD thesis, Department of Computer Science, University of Copenhagen (DIKU)(November 2011)).*

[14] *Lee uses a logic programming model and models a contract as a set of transition rules for a Petri net: Lee, R. M. (1988). A logic model for electronic contracting. Decision support systems, 4(1), 27-44*

[15] *See Prisacariu and Schneider's Contract Language (CL) that combines deontic, dynamic and temporal logics: Prisacariu, C., & Schneider, G. (2007). A formal language for electronic contracts. In International Conference on Formal Methods for Open Object-Based Distributed Systems (pp. 174-189), Springer, and Pace, G., Prisacariu, C., & Schneider, G. (2007). Model checking contracts–a case study. In International Symposium on Automated Technology for Verification and Analysis (pp. 82-97). Springer.*

[16] *The Business Contract Language (BCL) monitors contract events: Governatori, G., & Milosevic, Z. (2006). A formal analysis of a business contract language. International Journal of Cooperative Information Systems, 15(04), 659-685.*

[17] *Flood and Goodenough specify a contract as a deterministic finite automaton with transition rules (this may be thought of as a 'virtual machine' used as an executable specification): Flood, M. D., & Goodenough, O. R. (2015). Contract as automaton: the computational representation of financial agreements. Office of Financial Research Working Paper, (15-04). This has some resemblance to Lee's Petri net transitions (op. cit. Footnote 14) and to the finite state machine representation used by Molina-Jimenez et al: Molina-Jimenez, C., Shrivastava, S., Solaiman, E., & Warne, J. (2004). Run-time monitoring and enforcement of electronic contracts. Electronic Commerce Research and Applications, 3(2), 108-125.*

[18] *Also see: Prakken, H., & Sartor, G. (2002). The role of logic in computational models of legal argument: a critical survey. In Computational logic: Logic programming and beyond (pp. 342-381). Springer, Berlin, Heidelberg.*

A **domain-specific modelling language** (DSML) may be used as a specification language to draft a new contract or to create a formal model of a contract (as above, typically either to assist analysis of properties of the contract – such as consistency analysis or "what if" analysis – or to assist generation of code at a lower level in the language stack): these DSMLs may be customised to the drafting or modelling of contracts in a given business sector and are either (i) embedded in a general programming language (more understandable for a programmer) or (ii) designed as a separate language (more understandable for a lawyer, and perhaps using a **controlled natural language**), which may assist validation that the code is faithful to the agreement as described above.[21] Occasionally, a DSML may have a visual programming interface (specification with visual elements rather than text – see Section 4.2).

A **markup language** such as XML[22] can be used to annotate natural language with tags to provide additional information relating to presentation, structure or semantics and may act as a specification language, albeit in a limited way since its purpose is to annotate existing natural language expressions rather than to provide an alternative expression.[23]

- **Programming language.** Computer programmers construct software using a programming language. Often these are general-purpose languages, but **domain specific programming languages** (DSPLs) are designed for a specific purpose – for example, to target a specific application, or a specific distributed ledger platform.[24] DSMLs and DSPLs belong to the general category of Domain Specific Languages (DSLs). There are many programming languages, with differing styles of expression and with differing degrees of formality in terms of their defined syntax and semantics. The semantics (or meaning) of programming languages is at the heart of our understanding of how a computer program behaves and what it computes, and is an important tool for the validation of all but the simplest code.[25] Declarative languages such as the functional languages Haskell[26,27] and Miranda[28,29], and the logic language Prolog[30,31], have formally defined semantics and are especially helpful in providing certainty about the meaning of code; they may also be used either as a programming language or as a specification language or both.[32] Where there is a desire for flexibility in targeting multiple platforms, a programmer might construct software using a general-purpose programming language and then automatically (or semi-automatically) translate the software to a platform-specific programming language.

---

[19] *If a specification language is not itself a programming language (e.g. if it is a non-executable mathematical formalism) then it may be amenable to semantics-preserving translation into a programming language.*
[20] *If the natural-language contract makes use of a standard ontology (such as FIBO – op. cit. Footnote 10) then it will assist checking of the specification if the specification language is also able to use the terms from that ontology, with the same definitions.*
[21] *Embedded and non-embedded DSMLs are sometimes known as 'internal' and 'external' DSMLs.*
[22] *https://www.w3.org/TR/REC-xml/*
[23] *A markup language is not a programming language, but it can guide an application that views, edits or analyses text, and it can for example annotate an area of text to indicate that it is source code written in a programming language.*

- **Assembly language and object code.** Software written in a programming language must be converted to a lower-level language for execution by a computer. For example, a 'compiler' converts a program expressed in a programming language into a program expressed in an executable language: 'machine code' (see below). However, the compiler may employ several intermediate steps using intermediate languages: for example, **assembly language** is a lower-level language and can be further converted into executable machine code (see below) by an 'assembler', and **object code** is an intermediate form of machine code that needs to be 'linked' with other object code before it can be executed by a computer.

- **Executable languages, machine code and instruction sets.** Software is primarily created with the expectation that it will be executed on computer hardware. The final step is (almost) always that an electronic component fetches code as a package of binary digits (bits) from a storage medium, performs whatever action is defined by the code bits (which may include fetching and operating on data bits), and then fetches the next package of code bits. The bits may be structured into packages of 8 (a 'byte') or 32 (a 32-bit 'word') or larger. Both code and data in a modern digital computer are represented as sequences of packages of bits; what makes code different to data is only (i) it is stored where code is expected to be found, and (ii) (hopefully) it represents a sequence of valid actions. What action is defined by a particular package of bits is highly specific to the hardware (each understands a different **executable language**, often called its **instruction set**). A sequence of such bit patterns is called **binary code**, **machine code** or **native code**.

---

[24] *Examples of domain specific programming languages for code running on specific distributed ledger platforms include Solidity for Ethereum (https://en.wikipedia.org/wiki/Solidity), Plutus for Cardano (https://cardanoprogramming.com/cardano-plutus-programming-hello-world/) and Pact for Kadena (https://d31d887a-c1e0-47c2-aa51-c69f9f998b07.filesusr.com/ugd/86a16f_442a542b64554cb2a4c1ae7f528ce4c3.pdf). By contrast the domain specific programming languages CSL by Deon Digital (https://deondigital.com/docs/v0.38.0/) and DAML by Digital Asset (https://daml.com/) are not specific to any particular technology platform.*

[25] *The semantics of programming languages typically refers either to denotational or operational semantics: the former maps programming expressions (and by extension the whole program) to formal mathematical objects, whereas the latter creates proofs from logical statements about the way a program operates. As an example of the former see Scott, D. and Christopher Strachey, C. (1971)* Toward a mathematical semantics for computer languages *Oxford Programming Research Group Technical Monograph. PRG-6.*

[26] *Jones, S. P. (Ed.). (2003). Haskell 98 language and libraries: the revised report. Cambridge University Press.*

[27] *Thompson, S. (1999). Haskell: The Craft of Functional Programming. Addison-Welsey. Reading.*

[28] *Turner, D. (1986). An overview of Miranda. ACM Sigplan Notices, 21(12), 158-166.*

[29] *Clack, C., Myers, C., & Poon, E. (1995). Programming with Miranda. Prentice Hall.*

[30] *Warren, D. H., Pereira, L. M., & Pereira, F. (1977). Prolog-the language and its implementation compared with Lisp. ACM SIGPLAN Notices, 12(8), 109-115.*

[31] *Mellish, C. S., & Clocksin, W. F. (1981). Programming in PROLOG. Springer.*

[32] *See for example Turner, D. A. (1984). Functional programs as executable specifications. Philosophical Transactions of the Royal Society of London. Series A, Mathematical and Physical Sciences, 312(1522), 363-388.*

- **Runtime systems, virtual machines, interpreters and byte code.** Part of the executable code run by a computer is standard code inserted by the compiler to deal with low-level matters such as the layout and reuse of memory, the passing of arguments to (and results from) functions, and interfacing with the operating system. This is called the **runtime system** and its behaviour is an important part of the behaviour of a program that is converted to an executable language (see above). In order to validate whether code is faithful to the agreement, it is essential to understand how the runtime system operates because it may alter the semantics of the program (e.g. what it calculates and how it operates).

   A sophisticated runtime system may act as a **virtual machine**, providing a more complex instruction set than the underlying hardware; in this case a compiler would produce executable code in the language of that virtual machine and the behaviour of the virtual machine would be another vital component in establishing the validity of the code. A sophisticated virtual machine may define its own executable language at a high level of abstraction, similar to a programming language;[33] this is conceptually similar to an **interpreter** that reads a programming language in small portions and creates and runs executable code for each portion before processing the next. Some benefits of both compilation and interpretation may be obtained by partially compiling a program from a programming language into an executable form called **byte code**, that is subsequently processed by an interpreter. Byte code can be portable across different hardware platforms, each with its own interpreter for the byte code, where each interpreter may contribute its own bias (in terms of errors or mis-interpretations) to the code.

In summary, an understanding of the language stack is vital to understanding the complexity of validating whether smart contract code is faithful to the agreement. The computer hardware does not understand code written at any other than the lowest layer of the language stack, and considerable expertise in computer science (including, for distributed ledger implementations, the semantics of distributed systems), is required to ensure effective validation. However, a greater potential for error arises from the translation from natural language to programming language, which requires considerable expertise and experience in multiple domains such as computer science, law and the business sector. The issue of semantics is explored further in Section 3.

---

[33] *The concept of a 'virtual machine' may also be helpful in reasoning about the meaning of programs and translations between languages.*

## 2 *Natural and Formal Expression*

Given the desire to automate aspects[34] of a contract, it is necessary to know the meaning of the agreement and of those aspects that require automation. To be clear, this necessarily and fundamentally entails an analysis of semantics (see Section 3) *ante hoc*, and requires amongst other things an analysis of the natural language expressions contained in the written contract.

The writing of computer code to automate an agreement requires the anticipation of a range of possible events that may occur during the performance of the agreement. It does not necessarily require *all* possible events to be anticipated: typically the code will contain a default action to perform (such as to alert one or both counterparties) if an unanticipated event were to occur. It will also be necessary to anticipate possible sequences of events occurring in different orders, and to establish the semantics of the agreement in each of these contexts. Again, not all sequences need to be anticipated (just as the text of legal documentation does not anticipate all possible futures). When we talk of the semantics of an agreement we mean a formal description that includes (*inter alia*) those actions that parties should undertake and the changes that should occur to deontic aspects such as rights, permissions and obligations as a result of each of the anticipated sequences of events.

The ease with which such semantic analysis can be undertaken for a specific agreement may inform and constrain the choice of which aspects of an agreement are amenable to automation.

### 2.1 From contract to code

We focus on written agreements, where a large part of the agreement is expressed in writing. The legal text is typically not sufficient, but is an important component and it is necessary to investigate the natural language contract (both as a whole and in terms of the various textual components), to determine meaning and from that meaning to produce computer code.

The ease with which (aspects of) an agreement can be converted into code depends on how the contract is written, and in particular whether it is written using an uncontrolled natural language or a controlled natural language:

---

[34] *The selection of which aspects to automate has been covered for specific contracts elsewhere. See for example Clack, C. D., & McGonagle, C. (2019). Smart Derivatives Contracts: the ISDA Master Agreement and the automation of payments and deliveries. arXiv preprint arXiv:1904.01461 and also ISDA, King & Wood Mallesons: Smart derivatives contracts: From concept to construction (2018), available at: https://www.isda.org/a/cHvEE/Smart-Derivatives-Contracts-From-Concept-to-Construction-Oct-2018.pdf.*

**Uncontrolled natural language.** In this case the text generally has insufficiently constrained structure of syntax and semantics for automatic conversion into a programming or specification language, and the conversion must therefore normally proceed manually. This process may require expertise from multiple disciplines – primarily law and computer science with input from the relevant business sector, but also linguistics and logic – and this human involvement, coupled with the potential for miscommunication and misunderstanding between subject experts, is a potential source of error and must be carefully managed.

**Controlled natural language (CNL).** A CNL may retain the nuance and flexibility of a natural language and yet also be sufficiently structured for easier conversion to a programming or specification language. Current research is addressing the design of a CNL that is also a specification language (a technique known as 'Computable Contracts': see Section 4), thereby reducing the potential for human error during the conversion process.

In both cases (though more prevalent with an uncontrolled natural language) there will be an issue relating to whether (and if so how) automation of separate aspects of a contract should be brought together into a single computer program. The meaning of some aspects of a contract may convert easily, most notably where they do not depend on operational context: for example, a definition of a bank account number is straightforward to convert into code,[35] and the calculation of an amount to be paid may be no more than a description in natural language of an algebraic expression. Such aspects could be converted separately, but how should they be brought together into a single computer program such that (for example) calculations and payments are performed at the correct time? Furthermore, in general terms the substance to be converted will be more difficult than an algebraic expression: the meaning of each aspect (and therefore the code to be written) may not be self-contained due to interactions between clauses, and the text may not be representative of the entire agreement. Attention must also be given to the ways in which applicable law applies to the contract. In linguistics there is a helpful distinction between the semantics of the written text and the pragmatics of unwritten understanding between the writer and the reader of the text – in a written contract, the action of applicable law is part of the linguistic pragmatics. For example, implied terms may act to constrain the written terms of a contract, and in doing so they may change the rights, obligations, permissions or prohibitions that apply to the parties and this in term affects the deontic semantics of the agreement. This is of great importance to the conversion into smart contract code, and to the validation of that code, since both depend on a correct understanding of the semantics of the agreement.

An alternative to bringing the parts together into a single program might be to implement the code as a collection of autonomous and asynchronous programs communicating via the passing of electronic messages. However, validating the semantics of such a complex distributed implementation may be challenging (see Section 3.3).

---

[35] *Smart contract code running on a blockchain should also support the changing of such basic data during the term of a long-running agreement – this may conflict with the notion of "immutability" for some blockchain architectures, but it is an important design requirement.*

**2.2 Internal or external model**

If the parties agree on the smart contract code that will automate aspects of an agreement, they will almost always base that agreement on code written in a human-readable specification language or programming language (source code). Bearing in mind that in English law "The parties' contractual obligations may be defined by computer code",[36] then there are two possibilities that have implications for validation:

- The source code implements aspects of the agreement, but does not define any aspect of the agreement. This is the 'external model' defined by ISDA and Linklaters.[37]

- The agreement is defined in whole or in part by the source code (see above). This is the 'internal model' defined by ISDA and Linklaters.[38]

In the case of the external model there is a risk of human error in converting from a natural language to a specification or programming language. By contrast, in the case of the internal model some of the risk of human error during conversion from a natural language to a specification or programming language is removed, but the ability to express the substance of the agreement using a specification or programming language depends on how well the language fits the task of drafting a contract. For example, is it sufficiently flexible? and is it capable of expressing the contractual obligations that the lawyer wishes to express? It also depends on to the extent to which the different languages interface well with each other (e.g. can a clause written in a natural language easily and precisely make reference to an object defined in a programming or specification language? and *vice versa*?).

**2.3 Practical aspects of validating changes in code or agreement**

Both code and agreement may be subject to change (to illustrate the former, code may require updating due to a change in computer hardware or operating system, or because it is exhibiting anomalous behaviour), and a change in one will most likely require a change to the other. Where changes occur, validation must be repeated to ensure that the code remains faithful to the agreement. There are several optional strategies for where the source code is stored,[39] none of which make any legal difference but each of which can have practical consequences for the validation of changes in code or contract. For example:

---

[36] *UK Jurisidiction Taskforce (2019). Legal statement on cryptoassets and smart contracts. The LawTech Delivery Panel, Available: https://technation. io/about-us/lawtech-panel.*
[37] *ISDA & Linklaters (2017) Smart contracts and distributed ledger – a legal perspective: http://www.isda.org/a/6EKDE/smart-contractsanddistributed-ledger-a-legal-perspective.pdf*
[38] *Ibid.*
[39] *Where the executable form of the code is stored is of less interest for this discussion, though it is important operationally and for example Ricardian Contracts require bidirectional links between contract and executable form. See Grigg, I. (no date) The Ricardian Contract, http://iang.org/papers/ricardian_contract.html*

- **Strategy 1. Keep the contract and source code separate.** If for example the source code were deemed to implement rather than define the contract, this may be tempting due to the operational pragmatics of conversion and management. However, keeping the contract and code separate can be problematic in terms of version synchronisation when the agreement or code is revised – in practice, in the context of an organisation that manages very many agreements, the link between agreement version and code version may be lost, so that it may become unclear which version of the code relates to which version of the agreement.

- **Strategy 2. Attach the source code as an appendix to the contract.** This is a better strategy, and is especially helpful if for example the source code were deemed to be part of the contract. Version control for repeated update and validation are substantially easier. If required, the legal text may be annotated with references to lines of code in the appendix.

- **Strategy 3. Distribute the source code throughout the contract.** This has the advantage of placing lines of code visually adjacent to the legal text whose substance they are intended to automate (perhaps using a markup language to distinguish between code and non-code). There are well-established operational advantages to this 'literate' style of layout that mixes textual expressions and code.[40] If the only automation being performed is the writing of small amounts of code for extremely simple calculations or actions, it may be straightforward to include all of the smart contract code as multiple small additions to the legal text (though combining these small pieces into a single program may be complex, as mentioned previously). A more likely scenario is the combination of this strategy with either Strategy 1 or Strategy 2, where most of the source code is separate or in an appendix and a small amount of source code (or carefully structured text) is embedded in the contract to indicate either values for defined names[41] or specific arithmetic formulae to be used.[42,43]

- **Strategy 4. Do nothing – the contract is the source code.** It is possible for a contract to be expressed entirely in source code:
  - trivially, this is true because source code can contain textual data objects of any complexity (and the code could therefore simply define a text object to contain the entire natural language contract);

---

[40] *See Knuth, D. E. (1984). Literate programming. The Computer Journal, 27(2), 97-111.*

[41] *In the terminology of Ricardian Contracts, these names and values would be 'parameters' – part of the Ricardian Triple of prose, parameters and code: http://financialcryptography.com/mt/archives/001556.html*

[42] *This is also the original basis of Smart Contract Templates. See https://vimeo.com/168844103, op. cit. Footnote 3, op. cit. Footnote 74 and Clack, C.D. (2018) Smart Contract Templates: legal semantics and code validation, Journal of Digital Banking 2(4):338-352.*

[43] *Further examples of the use of parameters embedded in legal prose can be found in Hazard, J., & Haapio, H. (2017). Wise contracts: smart contracts that work for people and machines. In Trends and communities of legal informatics. Proceedings of the 20th international legal informatics symposium IRIS (pp. 425-432).*

- more usefully, because a specification language (for example) could: express deontic aspects[44] such as rights, obligations, permissions and prohibitions (and the smart contract code could monitor those deontic aspects during performance of the agreement); express operational aspects (actions) and temporal constraints; and record definitions (such as choice of jurisdiction) for later use (for example to be made available during dispute resolution, which itself might be automated); and so on.

If a specification language or programming language (that might also be a CNL) were used to define an entire contract, then we could say that "the contract is the source code" and there is no separation. This makes it considerably easier to manage validation following a change. It should however be noted that the contract is not the entire expression of the agreement, and validation must still be undertaken to ensure that the code is faithful to the agreement as a whole. See also Section 4.

## 3. Semantics

Several fundamental semantic issues arise when attempting to automate the performance of all but the simplest aspects of an agreement – especially high-value agreements where absolute certainty is required that the code is faithful to the agreement. For example, the parties may wish not only to automate the performance of certain actions at specified times and perhaps conditional on certain events, but also to automate the monitoring of deontic aspects that may themselves be conditional on certain events. For example, the parties may wish to automate the monitoring of actions at the time each action is performed to ensure that such action is supported by an obligation or a right and does not conflict with a prohibition. Furthermore, for example, if a contract were to contemplate multiple possible futures with a branching structure in time then certain sequences of actions might be permitted along one future time path but not along a different future time path: in this example, the parties might wish smart contract code to be generated to automate adherence to these temporal aspects.

Many observers have highlighted the need for verification and validation of smart contract code,[45,46,47,48] and verification alone is likely to be insufficient. Consider that the contract is drafted by lawyers, whereas the code is created by programmers: the lawyers often do not understand the code (which requires specialist knowledge in computer science) and the programmers often do not understand the contract (which requires specialist knowledge in law).

---

[44] *Here we use the term 'deontic' in the sense of formal deontic logic (see Von Wright, G. H. (1951). Deontic logic. Mind, 60(237), 1-15), where for example rights, obligations, permissions and prohibitions are addressed separately to operational issues (actions) and temporal issues (time). Although deontic expressions often denote obligations etc. with respect to actions, they operate at a meta-level: they may refer to an action but they are not the action itself.*
[45] *Al Khalil, F., Ceci, M., OBrien, L., Butler, T. (2017) A solution for the problems of translation and transparency in smart contracts. Tech. rep., Government Risk and Compliance Technology Centre, at: http://www.grctc.com/wp-content/uploads/2017/06/GRCTC-Smart-Contracts-White-Paper-2017.pdf.*
[46] *Harley, B. (2017) Are Smart Contracts Contracts? Tech. rep., Clifford Chance, at: https://www.cliffordchance.com/briefings/2017/08/aresmartcontractscontracts.html*
[47] *ISDA & Linklaters (2017) Smart Contracts and Distributed Ledger − A Legal Perspective: http://www2.isda.org/attachment/OTU3MQ==/Smart Contracts and Distributed Ledger A Legal Perspective.pdf*
[48] *Magazzeni, D., McBurney, P., Nash, W. (2017) Validation and verification of smart con-tracts: a research agenda. IEEE Computer Journal50(9), 50–57, Special Issue on Blockchain Technology for Finance.*

Lawyers and computer scientists are highly trained, analytical problem-solvers; as a result there is a high probability that with a small exposure to concepts of law a programmer might incorrectly believe that he or she understands the agreement, and similarly with a small exposure to concepts of computing a lawyer might incorrectly believe that he or she understands the code. Meetings between lawyers and programmers to help clarify the meaning of agreement and code may be unsuccessful (despite a surface appearance of understanding) because lawyers and programmers have substantially different perspectives and do not share the same language. For example:

- Both programmers and lawyers rely on a large corpus of knowledge that will not be expressed in the contract or the code (e.g. of how law applies to legal text, and of how source code is manipulated and processed by a compiler and a runtime system).

- Both lawyers and computer scientists use commonplace words as terms of art with deeply specialised meanings that require training to understand fully. A few simple examples will make the point:

    o 'execution' has a lay meaning (e.g. performing a sentence of death), a specialist meaning to a lawyer (e.g. signing a contract), and a specialist meaning to a programmer (e.g. the carrying-out by a computer of the instructions of a program).

    o 'performance' has more than one lay meaning (an act of presenting a form of entertainment, or the action of undertaking a task), a specialist meaning for lawyers (e.g. undertaking contractual obligations), and a specialist meaning for computer scientists (e.g. the amount – or computational complexity, depending on context – of memory and time[49] used by a program).

    o 'interpretation' may generally refer to 'explaining the meaning of something' or for a lawyer 'the post-hoc semantic analysis of legal text', or for a computer scientist 'on-the-fly creation and running of executable code from source code'.

    o 'construction' may generally refer to 'the act of building some real or abstract thing', or for a lawyer 'to determine the legal effect of a contract' or for a computer scientist 'creation of a computer program or formal model'.

- Computer scientists are accustomed to establishing semantics *a priori* and *ante hoc*, whereas lawyers are accustomed to semantic analysis (such as interpretation and construction) being deferred until a dispute occurs, *post hoc*, with potentially different outcomes depending on what semantics are inferred from the words and clauses of the contract.

---

[49] Where 'time' may be measured as a number of 'clock cycles' of a computer's CPU (the central processing unit).

The advantage of *post hoc* analysis is that it can focus on events that actually happened, rather than anticipating what the agreement would mean following many possible sequences of events that might happen. However, such *post hoc* semantic analysis is problematic for Smart Contracts, since it is difficult to write code to automate the performance of an agreement without knowing in advance what needs to be done – i.e. the semantic definition of what the agreement means must be undertaken *ante hoc*, and it is the *ante hoc* semantics that drive both the creation and the validation of the code. It is of course often true that the meaning of an agreement is not fully specified (because future events are not entirely knowable), but a small percentage of unknowns can be managed – computer scientists are accustomed to working with incomplete specifications (for example, for an unforeseen event the code could raise an alert with the parties and pause performance until instructed how to proceed).[50]

- Programmers are similarly accustomed to temporal aspects being well-defined with clear logic (including, for example, whether time is viewed as discrete or continuous), whereas lawyers are accustomed to highly flexible and context-sensitive interpretations of time (e.g. "the payment may be late, but is it *materially* late?" and "regardless of the actual time of an event it is *deemed* to have occurred at the start of the day"). Thus, decisions based on the timing of events may need to be made with a varying degree of precision, and the amount of variation in precision may not be known until the point at which the decision must be made (and sometimes not until after a decision is made).

With such deep differences in perspective and language, attempting to create and then validate smart contract code by asking programmers to view a written specification produced by lawyers is likely to be problematic. The specification may be misleading, confusing or unintelligible to the programmers, or may have multiple possible meanings thereby presenting the risk that the programmers might make an incorrect choice of meaning. Until the gap in linguistic semantics and pragmatics between lawyers and computer scientists is resolved, the creation and the validation of smart contract code for high-value or safety-critical agreements should ideally be undertaken by a multi-disciplinary team of lawyers and computer scientists working together closely:

---

[50] *The two-layer approach of 'interpretation' and 'construction' highlights an interesting issue: where they differ, which should guide the creation and validation of smart contract code? Another way to view this is to ask whether the parties wish to automate aspects aspects that may not be upheld by a court of law as well as aspects that would be upheld by a court of law?*

- first, to establish an agreed, detailed and unambiguous formal specification of the meaning of the agreement, and an identification of those parts of the agreement that the parties wish to automate;

- second, to agree on the technology platform (e.g. which Distributed Ledger Technology platform will be used);

- third, to collaborate in the creation and validation of the smart contract code.

If the first step were to prove difficult, then a practical way forward might be to adopt a 'rapid-prototyping' approach, where the first, second and third steps would be repeated for a sequence of prototypes of increasing complexity.

## 3.1 Semantics and Validation

In general terms, possible strategies for validating smart contract code (in the sense previously discussed of whether the code correctly automates the aspects – such as calculations and actions – that need to be automated, no more and no less, whilst not conflicting with any aspect of the overall agreement) include the following.

- **Exhaustive validation.** For very simple Smart Contracts, 'path testing' could be used, whereby every possible path through the agreement (for all possible contemplated future events, in all possible orders) and every possible path through the smart contract code would be validated. For each possibility it would be necessary to know what the code should be doing with respect to the semantics of those aspects of the agreement that require automation. It would also be necessary to understand the semantics of the entire agreement, to confirm that the smart contract code never conflicts with the agreement. This is an extremely time-consuming approach, it may prove impossible to cover all possible paths, and it is unlikely to be feasible for complex Smart Contracts.

- **Formal methods.** Mathematicians, logicians and computer scientists have been studying and developing formal proofs of the correctness of code since the 1930s. Much work has been done in the area of formal verification, though formal methods can also be applied to validation. Two example techniques are model-checking[51] and theorem proving[52]: with model-checking, both the code and the agreement could be modelled as automata (based on semantic specifications of each) and these two automata could be compared to determine if the behaviour of the code conforms to that of the agreement; by contrast, with theorem-proving both the code and the agreement could be expressed as logical formulae, with inference rules being used to demonstrate a mapping from the former to the latter.[53]

---

[51] *See for example Baier, C., & Katoen, J. P. (2008). Principles of model checking. MIT press.*
[52] *See for example Fitting, M. (2012). First-order logic and automated theorem proving. Springer Science & Business Media.*
[53] *A simple example, in a different field, is given in Lampérière-Couffin, Sandrine, et al. (1999) "Formal validation of PLC programs: a survey." 1999 European Control Conference (ECC). IEEE.*

The advantages of formal methods are that (i) within the context of their assumptions and those logical statements they seek to prove they can give a very great deal of certainty about the result, and (ii) this is a mature area, having been developed over many decades. The rub is in knowing that the automata or logic in each case is a full representation of the code or agreement respectively.[54] For the code, this is made considerably easier if it is written using a declarative language.

- **Valid by design.** The burden of validation could be eased considerably if the potential for human error during conversion from agreement to code were removed. One solution would be for lawyers to draft contracts using a programming language or a domain specific language (DSL). The former is not impossible, but unlikely: the latter would be considerably more attractive if it could be made to resemble a natural language, and almost certainly this would be a controlled natural language (CNL). If a DSL could be designed that were also a CNL customised for the task of drafting contracts, and if that DSL/CNL hybrid were to have clearly defined semantics then much of the conversion from agreement to code could be automated, with proofs of correctness at each step of the conversion down the language stack to executable code, and a large part of the final code would be 'valid by design'. The final executable smart contract code would also include an appropriate runtime system and perhaps standard ancillary code, and the entire code would be validated against the agreement; this process would be substantially facilitated by the fact that the code had been automatically created from the contract. This is the aim of the methodology known as 'Computable Contracts', which is discussed further in Section 4.

All strategies require a better understanding of the semantics of both the agreement and the code, which are further discussed below.

**3.2 Semantics of the agreement**

The analysis of the semantics of an agreement must consider multiple aspects including deontic aspects (i.e. rights, obligations, permissions, prohibitions etc),[55] operational aspects (actions to be performed) and temporal aspects (relating to and reasoning about time). Each such aspect may be specified independently, or preferably in combination[56] and it is essential for the formal semantic specification itself to be validated to determine whether it correctly captures the meaning of the contract.

Section 2.1 introduced the observation that some portions of legal text (such as definitions and algebraic expressions) may be trivial to convert into an expression in a programming or specification language. This is because they have simple semantics. In general, however, the meaning of an isolated portion of legal text cannot necessarily be derived independently because there is nothing to constrain another clause from modifying the meaning of the clause being studied, and a reading of the entire contract may be required to identify conflicts with other clauses. Furthermore, as mentioned previously, the meaning of a contract cannot necessarily be derived from the text alone; it is necessary also to consider the broader agreement and the role of law and how it affects the contract.

---

[54] *Important work is underway with respect to the formal semantics of blockchain platforms, for example in providing a formal model of the Ethereum Virtual Machine (to be used as part of the formal model of the smart contract code). See Hirai, Y. (2017). Defining the ethereum virtual machine for interactive theorem provers. In International Conference on Financial Cryptography and Data Security (pp. 520-535). Springer, Cham.*
[55] *These are concerned with contractual deontics rather than deontological ethics.*
[56] *Lee, R. M. (1988). A logic model for electronic contracting. Decision support systems, 4(1), 27-44.*

A contract may include both ambiguity and vagueness, both of which hinder the determination of meaning. We use the term 'ambiguity' to mean either (i) the use of a term or phrase that is inherently ambiguous (such as in the phrase "*I bequeath all my black and white horses*", which might mean "*all my black horses and all my white horses*" or "*all my horses that are both black and white*"),[57] or (ii) a word or expression that the parties believe to be unambiguous but by which they understand differing meanings (perhaps due to a misunderstanding over what the term denotes, such as whether 'Paris' is in France or Texas, or what it connotes, for example whether 'chicken' refers to any bird of the subspecies Gallus gallus domesticus, or a young chicken suitable for frying?). A bigger problem is vagueness: here we use this term to refer to expressions that do not have a single meaning, nor are they ambiguous with a small number of meanings from which to choose, but rather they have an indeterminate, possibly deferred meaning. Sometimes vagueness is inherited from the language used to create legal text (such as the words 'reasonable', 'fair' and 'timely') and sometimes a clause may be deliberately vague because it is the only way to achieve agreement between the parties.

When defining the semantics of an agreement, either to guide the creation of smart contract code, or to be used during the validation of smart contract code, or to be used to analyse an agreement, it is necessary to consider multiple aspects of meaning: for example, the deontic, operational and temporal aspects. Whether undertaken separately or in combination, we call the result a 'semantic specification'; it is a view of the entire agreement in formal terms that is amenable to formal analysis and formal logic. It is important that this specification gives a view of the agreement as a whole, since there are semantic interactions between clauses – the meaning of one clause may be affected by another clause, and so simply gathering a set of separately-derived semantic definitions of a number of separate aspects of the agreement would not be sufficient. When creating a semantic specification, three issues arise:[58]

1. **The separability problem.** The temporal, deontic and operational aspects are closely intertwined and difficult to separate (both in the natural language text of written contracts and in the development of formal logics). Although programmers tend to focus on the logic of actions such as payments and deliveries (the 'operational' aspects), it is important that they should also consider the deontic aspects that function at a meta level to reason for example about whether a party has the right, permission or obligation to perform each such action.

---

[57] *This example was drawn to the author's attention by Paul Lewis (Linklaters).*
[58] *Clack, C.D. and Vanca, G. (2018) Temporal Aspects of Smart Contracts for Financial Derivatives, Lecture Notes in Computer Science 11247:339-355, Springer-Verlag.*

Relevant logics include von Wright's deontic logic[59], Rescher and Urquhart's temporal logic[60], Azzopardi et al's contract automata[61], von-Wright's logic of action[62] and Prisacariu and Schneider's action-based logic[63]. Computer scientists also draw a distinction between 'denotational' and 'operational' semantics of computer languages, which may carry over into their perspective of the semantics of legal agreements.

2. **The isomorphism problem**. The structure of the semantic specification may be substantially different to the structure of the legal documentation. For example, a single clause may be mapped to multiple parts of the semantic specification, and one part of the specification may be derived from multiple clauses. This may make it difficult for a lawyer to understand and validate the semantics.

3. **The canonical form problem.** There may be many different ways to structure the semantic specification for a given legal agreement; specifically, there may be no agreed standard way to structure the semantics such that (i) two contracts with the same meaning will always have structurally identical semantic specifications, and (ii) two contracts with different meaning will always have structurally different semantic specifications (such a standard structure is sometimes called a "canonical form"). This may make it difficult to compare specifications automatically to see if they have the same meaning, which may be helpful in a number of different circumstances such as:

    o determining whether a collection of changes to an agreement that are intended to be benign (e.g. to simplify the wording) has resulted in any change in meaning;

    o determining whether a collection of changes to an agreement that are intended to change the meaning has achieved the required change; and

    o determining whether different versions of a proposed collection of changes to an agreement, drafted by different lawyers, result in the same meaning.

---

[59] *von Wright, G.H. (1965). And Next. Acta Philosophica Fennica, Fasc. XVIII, Helsinki. pp. 293-301. von Wright, G.H. (1967) The Logic of Action4 Sketch. in N. Rescher, ed. The Logic of Decision and Action. pp.121-136. Pittsburgh: Univ. Pittsburgh Press. von Wright, G.H. (1968) An Essay in Deontic Logic and the General Theory of Action. Acta Philosophca Fennica, Fasc. XXI. Amsterdam: North Holland Publishing*
[60] *Rescher, N. and A. Urquhart (1971) Temporal Logic. Wien: Sprmger- Verlag*
[61] *Azzopardi S., Pace, G.J., Schapachnik F., and Schneider G. (2016) Contract automata. Artif Intell Law 24:203–243*
[62] *Von Wright, G.H. (1967) "The Logic of Action – A Sketch" in N. Rescher, ed. The Logic of Decision and Action, University of Pittsburgh Press, pp. 121-136.*
[63] *Prisacariu C., Schneider G. (2009) CL: An Action-Based Logic for Reasoning about Contracts. In Logic, Language, Information and Computation. WoLLIC 2009. Lecture Notes in Computer Science, vol 5514. Springer, Berlin, Heidelberg*

These are not the only issues that arise: e.g. Pace and Schneider explain other semantics-based challenges.[64]

## 3.3 Semantics of the code

Just as it is impossible to know what code to write if the meaning of the agreement has not been established, so too it is impossible to know whether the written code is faithful to the agreement if the meaning of the code is not known.

Determining the meaning of a computer program may be achieved in several ways. For example: by ascribing mathematical meanings to syntactic expressions (mapping syntactic expressions to mathematical objects),[65] by creating proofs from logical statements about the way a program operates, by defining how a syntactic expression would change the state of a defined virtual machine, or by describing axioms that apply to a syntactic expression. Formally-based languages such as Haskell and Prolog are generally more amenable to formal determination of semantics. These techniques are used for very complex, very-high-value, or safety-critical software because of the impracticality (sometimes impossibility) of validating all possible behaviours of a program. Such semantic analysis of programs has been an established technique for decades.

Various issues arise in determining the semantics of a program. For example, the meaning of an isolated portion of code from a computer program cannot necessarily be derived independently, and may need to be derived in context by reading the entire program in the order in which it was written. This may be especially true for imperative source code and less true for declarative source code. For example, declarative languages based on the λ-calculus benefit from the Church-Rosser property that the order in which evaluation rules are applied (if that order terminates) does not make a difference to the eventual result. Furthermore, it is necessary to consider not just the actions of the source code but also of the compiler and runtime system as explained below:

- Source code is often modified during the process of compilation in order to improve performance (such as execution speed and parsimonious use of computer memory).[66] In most cases the modification does not change the meaning of the code, but occasionally (especially for complex optimisations) the meaning may be slightly changed. Furthermore, it has been demonstrated[67] that many well-established, mature compilers sometimes produce executable code that may sometimes be incorrect (it does not exactly follow the source code instructions).[68]

---

[64] *Pace, G. J., & Schneider, G. (2009, February). Challenges in the specification of full contracts. In International Conference on Integrated Formal Methods (pp. 292-306). Springer, Berlin, Heidelberg.*
[65] *This deduces the meaning of a program just by "looking at the code", in a similar fashion to understanding what a mathematical expression means (what it will calculate) just by looking at it.*
[66] *For example, repeated calculation of the same constant expression is unnecessary and the code may be modified so that the calculation is only performed once.*
[67] *Contrary to Note 106 of op. cit. Footnote 36.*

- The meaning of the code may be more difficult to ascertain where the runtime system is structured as multiple autonomous communicating parts, since it is necessary to consider the numerous different ways in which the parts may communicate (though if structured appropriately the semantics may be tractable, for example via use of process calculi[69,70]).

In general terms, the lower-level implementation software (compilers, interpreters, optimisers, code generators, linkers, loaders, operating system, runtime system etc.) should not be assumed to be free from semantic distortion (i.e. unable to preserve the semantics of the source code). The preservation of semantics must be tested and demonstrated. If they are not so demonstrated, then the executable code that runs on a computer may not perform in the way predicted by the semantics of the source code at the top of the language stack. For example, a semantic specification of a program written in a language near the top of the language stack may give confidence about the behaviour of the code with respect to an intermediate level description of a virtual machine, but the implementation of that virtual machine on a specific technology platform must itself be tested for the existence of obvious coding faults and validated for the existence of more subtle errors that may change the semantics of the code. As an example of a semantics-changing error, consider "data races" where one operation may speed ahead of another that started earlier, thereby leading to a change in the sequencing of actions. Mature technology tends to be fairly stable in this regard, but less mature technology (such as new technology platforms, and new platform-specific languages) may be less stable.

It appears to be inescapable that, especially for contracts of very high value, the executable code must always be thoroughly tested and validated for fidelity to the agreement (as previously explained). This would be especially true, for example, if the 'internal model' were used and if the parties were legally bound by what the code does, rather than by what it says.[71] However, the use of formally-based languages with known semantics at the higher levels of the language stack can substantially reduce the burden of validation.[72]

---

[68] *Interesting examples can be found in Yang, X., Chen, Y., Eide, E., & Regehr, J. (2011). Finding and understanding bugs in C compilers. In Proceedings of the 32nd ACM SIGPLAN conference on Programming language design and implementation (pp. 283-294).*
[69] *Hoare, C. A. R. (1978). Communicating sequential processes. Communications of the ACM, 21(8), 666-677.*
[70] *Milner, R. (1980). A calculus of communicating systems. Springer Verlag, ISBN 0-387-10235-3.*
[71] *See Paragraph 165 of op. cit. Footnote 36.*
[72] *An example of this for a mission-critical financial system is given in: Braine, L., Haviland, K., Smith-Jaynes, O., Vautier, A. & Clack, C. (1998). Simulating an object-oriented financial system in a functional language. https://arxiv.org/abs/2011.11593*

## *4. Computable Contracts*

As long as the agreement and the code remain separate there will be a need to validate whether the code is faithful to the agreement, and it has been demonstrated in the foregoing discussion that this is problematic. The task includes the following five steps:

1. for that part of the agreement to be automated, determining *ante hoc* what it means (so that we know what the code should do) – this is not necessarily straightforward;

2. for the agreement as a whole, determining *ante hoc* what it means (so that we can check that the code does not conflict with the agreement as a whole) – this is the semantic specification of the agreement, and is not necessarily straightforward;

3. for the written source code, determining *ante hoc* what it means (what will the source code do, for all specified inputs and circumstances?) – this is the semantic specification of the source code and may be more or less easy to achieve depending on the chosen language;

4. determining whether the source code, as expressed in its semantic specification, correctly automates the meaning of the selected aspects of the agreement, whilst not conflicting with the whole agreement (as expressed in the semantic specification of the agreement);

5. determining whether the implementation software (comprising compilers, interpreters, optimisers, code generators, linkers, loaders, operating system, runtime system etc.) correctly implements the semantics of the source code.

An emerging methodology in the field of Smart Contracts is that of Computable Contracts,[73] where part or all of the written legal documentation is written in a constrained natural language (CNL) that is also a domain specific language (DSL), where the latter has formally defined syntax and semantics and can be converted into a programming language via semantics-preserving transformations. Thus, that part of the written legal documentation will serve as a single artefact expressing both contractual obligations and the automated implementation of those obligations – it would be both contract and code, understandable to humans (lawyers and programmers) and computers (e.g. it can be taken directly as input to a compiler, for automatic conversion into executable code).

With an appropriately designed DSL, a Computable Contract could also be (or be automatically converted into) a formal semantic specification: the single artefact could be contract, code and semantic specification.

---

[73] *The term 'computable contract' was first introduced by Surden. See Surden, H. (2012). Computable contracts. UCDL Rev., 46, 629.*

With a single artefact, the task of checking whether the meaning of the source code matches the meaning of the agreement would be much simpler, and validation would include the following steps:

1. for the agreement as a whole, determining *ante hoc* what it means – this is made easier by the fact that the the Computable Contract (the single artefact mentioned above) has a clear semantic specification;

2. determining whether the semantic specification of the Computable Contract correctly automates the meaning of the selected aspects of the agreement, whilst not conflicting with the whole agreement (as expressed in the semantic specification of the agreement);

3. determining whether the implementation software (compilers, interpreters, optimisers, code generators, linkers, loaders, operating system, runtime system etc.) correctly implements the semantics of the Computable Contract.

## 4.1 Markup languages and templates

One approach to linking a written contract with its associated smart contract code is to use a markup language to add annotations to the contract. With this approach, some parts of the written text are annotated to indicate that they are natural language text and other parts are annotated to indicate that they are parameters (e.g. named values) that communicate with the code. Although the written contract and code remain separate, they are at least linked, and the concept can be extended so that the programming language source code is contained within the same text file as the written contract. Further markup annotations can also be applied to the natural language text, for example to provide bilateral linkages between separate documents and to annotate clauses to indicate their purpose.[74,75]

Substantial work has been done using this approach. Early work by Grigg proposed the term 'Ricardian Contract' for financial trading, aiming to achieve Smart Contracts that are simultaneously understandable by humans and computers by use of a markup language and linkages between contract and code.[76,77] Smart Contract Templates[78] extended Ricardian Contracts and proposed an abstract specification for structuring contracts, together with requirements and principles for templating. Hazard and Haapio[79] extended both Ricardian Contracts and Smart Contract Templates with their work on encapsulated pieces of legal text (that they call 'prose objects') that may contain definitions of values for names that link to smart contract code.[80] The markup languages LegalXML[81] and LegalRuleML[82] provide, respectively, data schemas and a rule interchange language for the legal domain.[83] Other straightforward markup languages for contract drafting include OpenLaw[84] and the Accord project[85]: the latter for example provides a library of contract and clause templates and the template language Cicero (combined with the expression language Ergo) to bind declaratively any existing natural language text to a data model.[86] At the time of writing, a British Standards Institute Publicly Available Specification for a standardised approach to contract templates is in its public consultation phase.[87]

---

[74] See for example Clack, C. D., Bakshi, V. A., & Braine, L. (2016). Smart Contract Templates: essential requirements and design options. arXiv preprint arXiv:1612.04496
[75] Also see http://www.commonaccord.org/ and https://www.accordproject.org/
[76] Grigg, I. (2000). Financial cryptography in 7 layers. In International Conference on Financial Cryptography (pp. 332-348). Springer, Berlin, Heidelberg.
[77] Grigg, I. (2004). The ricardian contract. In Proceedings. First IEEE International Workshop on Electronic Contracting, 2004. (pp. 25-31). IEEE.
[78] Op. cit. Footnote 42 and op. cit. Footnote 74.
[79] Hazard, J., & Haapio, H. (2017). Wise contracts: smart contracts that work for people and machines. In Trends and communities of legal informatics. Proceedings of the 20th international legal informatics symposium IRIS (pp. 425-432).

The use of a markup language to provide templates for Smart Contracts is a pragmatic approach to co-ordinating the requirements and activities of (i) drafting legal contracts; (ii) integrating those contracts with computerised business processes; and (iii) managing smart contract code for the automation of (some aspects of) those contracts. However, although the use of markup languages is currently popular there are several problems with this approach:

- The semantics of the agreement are only known to the extent that they are expressed in the tags attached to the legal text (which could be only a very small extent) and only for those aspects of meaning for which a tag exists.

- The meaning of a clause may be too complex to express with a simple tag. Two obvious solutions to this are (i) giving simple semantic tags to simple textual elements and defining an overarching analysis to derive the semantics of each clause from the semantics of its elements, or (ii) developing a tagging language to support complex semantic tags. However, both of these solutions are at odds with the essential simplicity of markup languages.

- In most cases the code would still be written in a programming language that may be opaque to lawyers.

- Although the contract and the code may have been brought together physically, in most cases the code would remain logically separate from the contract (even if parts are interspersed throughout the clauses of a contract).

- A markup language only makes a small contribution to address the key problem of how to validate whether the code is faithful to the agreement.

---

[80] *CommonAccord provides a collection of sample prose objects: http://www.commonaccord.org/*
[81] *http://www.legalxml.org*
[82] *https://www.oasis-open.org/committees/tc_home.php?wg_abbrev=legalruleml*
[83] *LegalRuleML focuses on markup for legislation. Grosof and Poon provide a rule-based approach to representing contracts that builds on RuleML and process knowledge descriptions from Semantic Web ontologies: Grosof, B. N., & Poon, T. C. (2003). SweetDeal: representing agent contracts with exceptions using XML rules, ontologies, and process descriptions. In Proceedings of the 12th international conference on World Wide Web (pp. 340-349).*
[84] *https://www.openlaw.io/*
[85] *https://www.accordproject.org/*
[86] *https://github.com/accordproject/cicero*
[87] *https://standardsdevelopment.bsigroup.com/projects/2018-03267#/section*

## 4.2  Domain specific programming languages

The key vision of Computable Contracting is the use of a single artefact to express both the contractual obligations and the smart contract code. This goes much further than marking-up natural language text, and envisions the design of a new language for drafting contracts. A first step in this direction has been the use of Domain Specific Languages (DSLs).

For many decades DSLs have been proposed and used to help in the creation of legal documentation as well as to help with automating the performance of agreements. According to both the original Szabo definition and the Clack et al definition, the use of a DSL to support the automated performance of an agreement would make it a Smart Contract. There has also been substantial use of DSLs in a Computable Contract role, providing a single semantic specification of contract and code. A very brief history of DSLs in this context is given below, much of which has been in the realm of financial trading agreements.

Early work by van Deursen and colleagues[88] developed the domain-specific language 'RISLA' for designing interest rate financial products in a way that was easy for financial engineers to understand. The language (which could be automatically translated into the COBOL programming language) was upgraded with (i) a component library to improve modularisation; and (ii) a questionnaire style of user interface. Peyton Jones et al[89] subsequently used a compositional style of programming to model the core product definitions of financial contracts, proposed as a DSML but also with an example implementation as a DSPL embedded in the functional programming language Haskell. Andersen et al[90] extended the work of Peyton Jones et al to the exchange of money, goods and services amongst multiple parties and provided a formal representation of contracts that supports definition of user-defined contracts and user-definable analysis of their state before, during and after execution. Henglein et al[91] extended this further to demonstrate how formal contract specifications provide the core of a process-oriented event-driven architecture. Seijas and Thompson's domain-specific language 'Marlowe' is also based on the compositional style of Peyton Jones et al, extended to embrace issues that arise when executing financial contracts on distributed ledgers.[92] Marlowe is embedded in the programming language Haskell and has a formal semantics that supports analysis of Marlowe specifications of agreements (which they call 'contracts', and which can be used to specify Smart Contracts). Although 26not yet developed, Goodenough proposes the development of a Legal Specification Protocol that *"is independent of natural language"* – the early indications are that this might eventually be either a DSML or DSPL.[93]

---

[88] *van Deursen, A. (1994) Executable language definitions: case studies and origin tracking techniques, Ph.D. thesis, University of Amsterdam.  Arnold, B. R. T., van Deursen, A. and Res, M. (1995) An algebraic specification of a language for describing financial products, in ICSE-17 Workshop on Formal Methods Application in Software Engineering, IEEE Computer Society Press, pp. 6–13.  van Deursen, A. & Klint, P. (1998). Little languages: little maintenance?. Journal of Software Maintenance: Research and Practice, 10(2), 75-92.*
[89] *Peyton Jones, S., Eber, J. M., & Seward, J. (2000). Composing contracts: an adventure in financial engineering (functional pearl). ACM SIGPLAN Notices, 35(9), 280-292.*
[90] *Andersen, J., Elsborg, E., Henglein, F., Simonsen, J. G., & Stefansen, C. (2006). Compositional specification of commercial contracts. International Journal on Software Tools for Technology Transfer, 8(6), 485-516.*
[91] *Henglein, F., Larsen, K. F., Simonsen, J. G., & Stefansen, C. (2009). POETS: Process-oriented event-driven transaction systems. Journal of Logic and Algebraic Programming, 78(5), 381-401.*
[92] *Seijas, P. L., & Thompson, S. (2018). Marlowe: Financial contracts on blockchain. In International Symposium on Leveraging Applications of Formal Methods (pp. 356-375). Springer, Cham.*

Building on the compositional approach of functional programming languages, several commercial DSLs have been designed to help write smart legal contracts for a variety of different distributed ledgers: for example, the 'Contract Specification Language' (CSL) from Deon Digital,[94] and the 'Digital Asset Modeling Language' (DAML) from Digital Asset[95]. Both of these examples are used as DSMLs, yet the style is programmatic (i.e. "in the style of a programming language") so that they may also be designated as DSPLs.

Marlowe has an optional visual user interface (using an adaptation of the visual language Google Blockly[96]) that provides a different style of specification using visual blocks as a metaphor for modules. Skotnica and Pergl have similarly suggested a visual DSL for modelling smart contract code.[97] Morris has developed the logic-based visual language 'Blawx' for specifying legal contracts.[98] Whilst visual programming languages are often aimed at novice programmers, they can be used to extend expressibility of a language,[99] and more generally Haapio and colleagues have collaborated to combine concepts from visualisation and contracting.[100],[101]

DSPLs are powerful and can express many operational aspects (e.g. payments, deliveries, and business logic) of smart legal contracts, especially in the realm of financial contracts. The better DSPLs provide a semantic certainty and clarity (caveat the previously discussed problems that may occur lower in the language stack) that makes them well suited to defining elements of the legal agreement (the 'internal model'). However, there is not yet sufficient evidence of their ability to support the specification and automation of deontic aspects[102] with the required degree of subtlety for lawyers to draft high-value smart legal contracts. Nor is there yet an agreed methodology by which the smart legal contract should be structured and elements included in a standardised way. Finally, the style of textual expression used in the embedded DSPLs is very similar to computer programming and may not always be easily and immediately understood by (or be attractive to) lawyers, even for very simple specifications (Figure 1 provides illustrative examples of DSPL drafting styles).

---

[93] *Goodenough, O. (2019) Developing a Legal Specification Protocol: Technological considerations and requirements. CodeX white paper. https://law.stanford.edu/wp-content/uploads/2019/03/LSPWhitePaperJan1119v021419.pdf*
[94] *https://deondigital.com/docs/v0.38.0/*
[95] *https://daml.com/*
[96] *Fraser, N. (2014). Google blockly-a visual programming editor. URL: http://code. google. com/p/blockly.*
[97] *Skotnica, M., & Pergl, R. (2019). Das Contract-A Visual Domain Specific Language for Modeling Blockchain Smart Contracts. In Enterprise Engineering Working Conference (pp. 149-166). Springer, Cham.*
[98] *See https://www.blawx.com/ and Morris, J. (2021) Rules as Code: How Technology May change the Language in which Legislation is Written, and What it Might Mean for Lawyers of Tomorrow https://s3.amazonaws.com/us.inevent.files.general/6773/68248/1ac865f1698619047027fd22eddbba6e057e990e.pdf*
[99] *For example, Braine and Clack have demonstrated how a visual notation can be used to facilitate the integration of two very different programming styles: Braine, L., & Clack, C. (1997). Object-flow. In Proceedings. 1997 IEEE Symposium on Visual Languages (Cat. No. 97TB100180) (pp. 418-419). IEEE.*
[100] *Wong, M., Haapio, H., Deckers, S., & Dhir, S. (2015). Computational contract collaboration and construction. In Co-operation. Proceedings of the 18th International Legal Informatics Symposium IRIS (pp. 505-512)*
[101] *Haapio, H., Plewe, D., & deRooy, R. (2016). Next generation deal design: comics and visual platforms for contracting. In Networks. Proceedings of the 19th International Legal Informatics Symposium IRIS (pp. 373-380)*
[102] *As previously observed in Footnote 44, the right to do something doesn't necessarily entail an action (the party may never exercise that right). Thus the right (deontic aspect) acts at a meta-level over the action (operational aspect). Rights, permissions and prohibitions can be expressed conditionally and therefore vary dynamically during the performance of the contract, and the automation of deontic aspects tends to lead to automation of the monitoring of performance, in addition to automating the performance itself.*

```
type CakeOrder: Event {
  amount: Int,
  receiver: Agent,
  item: String
}
type CakeDelivery: Event {
  receiver: Agent,
  item: string
}
template entrypoint CakeSale0(customer,shop,amount,item)=
  <buyer> order: CakeOrder
  where
    order.amount = amount &&
    order.receiver = shop &&
    order.item = item
  then
  <seller> delivery: CakeDelivery
  where
    delivery.receiver = customer &&
    delivery.item = item
```

```
escrow :: Contract
escrow = CommitCash
            iCC11 (ConstMoney 450) 10 100
            (When (OrObs (two_chose alice bob carol 0)
                         (two_chose alice bob carol 1))
                  90
                  (Choice (two_chose alice bob carol 1)
                          (Pay iP1 alice bob
                               (AvailableMoney iCC1) 100
                               redeem_original)
                          redeem_original)
            Null
```

**Figure 1: Drafting style in CSL (left)[103] and Marlowe (right)[104].**

### 4.3 Controlled natural language

With a CNL that is also a DSL (i.e. where the top two layers of the language stack have been merged), it is envisaged that drafting lawyers could ensure consistency of expression and structure for a written contract, without needing to become programmers. The combined CNL/DSL would, for example, be a structured variant of English and the process of drafting a contract would use English vocabulary and sentence construction – however, the CNL/DSL would also have sufficiently well defined syntax and semantics to enable automatic conversion to lower layers in the language stack (for those aspects requiring automation), culminating in executable smart contract code to control a computer.[105]

For Smart Contracts this would remove the error-prone step of manual conversion from natural language to a specification or programming language. With help from a customised user interface, it could be impossible to write a contract that could not be automatically convertible to executable code. This would provide substantial advantages for validating smart contract code, whilst remembering that the code must be validated against the entire agreement, not just the written contract.

---

[103] Based on https://docs.deondigital.com/v0.60.0/src/guidechapters/yourfirstcontract.html
[104] Based on Seijas, P. L., & Thompson, S. (2018, November). Marlowe: Financial contracts on blockchain. In *International Symposium on Leveraging Applications of Formal Methods* (pp. 356-375). Springer, Cham.
[105] Furthermore, it is likely that the process of drafting using a CNL would utilise a customised word processor with an advanced user interface (textual, visual, or a combination of both) to guide correct usage. An example of such an application is Juro (https://juro.com/). Similar tools (e.g. structure editors and source-code editors: https://en.wikipedia.org/wiki/Structure_editor and https://en.wikipedia.org/wiki/Source-code_editor) have long been available to computer programmers.

Advantages would also accrue, in some sectors such as financial services, from the standardisation of contract language and semantics.[106] Further advantages with respect to automated analysis of contracts (e.g. to highlight conflicts between clauses, missing clauses or incomplete expressions) include *"helping a practitioner clarify what is going on, even without encoding those statements into software"* [107] and may potentially have impact more generally in legal drafting.[108]

Attempto Controlled English (ACE)[109] is a mature CNL that generates Prolog code and maps onto the Web Ontology Language (OWL), though it has not yet been used directly in the specification of legal contracts.

The language 'L4' is currently under development and aims to be a combined CNL and DSL.[110] Similarly, Stanford University's CodeX centre proposes a computable contracting approach with the development of a Contract Description Language that is a single artefact that does not require programming yet is 'machine-understandable'.[111]

More immediately, Kowalski's 'Logical English'[112] combines a controlled natural language with a prototype implementation in the logic programming language Prolog. Kowalski's describes Logical English as being *"modelled on the language of law"* and *"designed not only to be understood without computer training, but to be useful for a wide range of computer applications, including legal applications involving smart contracts"*. Kowalski and his colleagues at Imperial College London have been working on the logical specification of legal language for decades[113,114] and this recent work builds on the Logic-Based Production System (LPS) programming language developed with Sadri and Calejo.[115] Logical English was developed with assistance from Davila and Karadotchev; in particular, Karadotchev's MSc dissertation provides a case study of using Logical English to express parts of the ISDA Master Agreement for financial transactions,[116] and further work by Kowalski and Datoo[117] has focused on the use of Logical English to standardise the wording of legal clauses concerning Automatic Early Termination in ISDA Master Agreements. Figure 2 illustrates the style of the current version of Logical English that can be translated directly into Prolog. Prolog itself is a declarative specification language that is suitable for automatic conversion to lower layers in the language stack and can therefore contribute to the generation of smart contract code.

---

[106] *The benefits of standardisation of contracts are well established in the financial sector (e.g. ISDA (2003) User's guide to the ISDA 2002 master agreement) and the construction sector (e.g. Chappell, D. (2007) Understanding JCT Standard Building Contracts. 8th ed. ISBN 978-0-415-41385-5) and are also being used in other sectors (Martin, K. (2018). Deconstructing contracts: contract analytics and contract standards. In Data-Driven Law (pp. 33-34). Auerbach Publications). See also ISDA's 'Clause Project' work on standardisation of language used in ISDA Schedules.*
[107] *Op. cit. Footnote 93.*
[108] *Cummins, J., & Clack, C. (2020). Transforming Commercial Contracts through Computable Contracting. arXiv preprint arXiv:2003.10400.*
[109] *Fuchs, N. E., Höfler, S., Kaljurand, K., Rinaldi, F., & Schneider, G. (2005). Attempto controlled english: A knowledge representation language readable by humans and machines. In Reasoning web (pp. 213-250). Springer, Berlin, Heidelberg.*
[110] *https://github.com/smucclaw/dsl*
[111] *http://compk.stanford.edu/*
[112] *Kowalski, R. (2020). Logical English. In proceedings: Logic and Practice of Programming (LPOP), http://www.doc.ic.ac.uk/~rak/papers/LPOP.pdf.*
[113] *Kowalski, R. A. (1984). Logic for knowledge representation. In International Conference on Foundations of Software Technology and Theoretical Computer Science (pp. 1-12), Springer.*
[114] *Sergot, M. J., Sadri, F., Kowalski, R. A., Kriwaczek, F., Hammond, P., & Cory, H. T. (1986). The British Nationality Act as a logic program. Communications of the ACM, 29(5), 370-386.*
[115] *Kowalski, R. A., Sadri, F., & Calejo, M. (2017). How to do it with LPS (Logic-Based Production System). In RuleML+ RR (Supplement). http://ceur-ws.org/Vol-1875/paper16.pdf*
[116] *Karadotchev, V. (2019) First steps towards Logical English, MSc dissertation, Imperial College London.*

```
It is not the case that
    it is an obligation that a party pays to a counterparty
    an amount in a currency for a transaction on a date
if  it is an obligation that the party pays to the counterparty
    a net amount in the currency for the transaction on the date

It is an obligation that a party pays to a counterparty
    a net amount in the currency for a transaction on a date
if  the net amount is a larger aggregate amount minus a smaller aggregate amount
and the larger aggregate amount is the sum of each amount of each payment by the
    party to the counterparty in the currency for the transaction on the date
and the smaller aggregate amount is the sum of each amount of each payment by the
    counterparty to the party in the currency for the transaction on the date
```

**Figure 2: example drafting style in Logical English. [118]**

From the example given in Figure 2 it can be seen that the current version of Logical English is much closer to natural language than the DSL examples given in Figure 1, and is somewhat closer to the computable contracting concept of a single artefact that is understandable to lawyers *and* understandable to computers.

The language Lexon[119] is also a combined CNL and DSL and specifically targets the generation of smart contract code for blockchains, aiming for automation of performance rather than analysis of contract semantics. Lexon comes with tools to support an extensible grammar, a formal syntax definition is being developed with a small core operational semantics, and the current version of the language can generate Solidity, Javascript or Sophia[120] as output. Figure 3 illustrates the drafting style for Lexon: notice that whereas the Logical English example in Figure 2 expresses deontic aspects (obligations), the Lexon example in Figure 3 expresses actions (since its purpose is to generate code to make the relevant payments).

---

[117] *Kowalski, R. and Datoo, A. (2020). Logical English meets Legal English for Swaps and Derivatives. http://www.doc.ic.ac.uk/~rak/papers/Logical%20English%20meets%20Legal%20English.pdf.*
[118] *Based on an example in: http://www.doc.ic.ac.uk/~rak/papers/Logical%20English.pdf.*
[119] *Diedrich, H. (2020). Lexon Bible: Hitchhiker's Guide to Digital Contracts. Wildfire Publishing. ISBN 978-1656262660. See also Idelberger's comparative analysis of Lexon: Idelberger, F. (2020) Merging Traditional Contracts (or Law) and (Smart) e-Contracts – a Novel Approach. In proceedings The 1st Workshop on Models of Legal Reasoning, Sao Paolo, Brazil, https://lawgorithm.com.br/wp-content/uploads/2020/09/MLR2020-Florian-Idelberger.pdf*
[120] *https://aeternity-sophia.readthedocs.io/en/latest/contracts/*

```
LEX Netted Payment #1.

"Party One" is a person.
"Party Two" is a person.
"Total Payable of Party One" is an amount.
"Total Payable of Party Two" is an amount.

CLAUSE: Register Payable By Party One.
The Total Payable of Party One is increased by a given Amount.

CLAUSE: Register Payable By Party Two.
The Total Payable of Party Two is increased by a given Amount.

CLAUSE: Daily Netting.
If the Total Payable of Party One is greater than the Total Payable of Party Two, then
Party One pays the Net Amount to Party Two.
If the Total Payable of Party Two is greater than the Total Payable of Party One, then
Party Two pays the Net Amount to Party One.
Afterwards, terminate the contract.

CLAUSE: Net Amount.
"Net Amount" is defined as the difference between
the Total Payable of Party One and the Total Payable of Party Two.
```

**Figure 3: example drafting style in Lexon[121]**

The approach of computable contracts is a specific methodology within the context of Smart Contracts. Computable contracts, if successful, have the potential to bring substantial benefit to the validation of smart contract code, and are a key component of current research and development in Smart Contracts. Research challenges for this methodology include:

- The representation of both (i) meaning that is known in advance; and (ii) meaning that is not fully known in advance and may require dynamic or post-hoc consideration of context and facts. The latter for example includes:

    o important legal phrases and words whose semantics are difficult to define (simple examples include the words 'reasonable', 'material' and 'timely'); and

    o deliberately vague clauses (as discussed in Section 3.2).

- The representation of points in the contract where human discretion is desired (and an understanding of how this will be implemented as interaction with the running code).[122]

- The representation of complex temporal aspects of a contract.[123]

- The investigation of new modular forms of expression for contracts, to enable greater encapsulation and re-usability of components (such as expressions and clauses). A relevant observation here is that with natural language contracts there is nothing to prevent two or more clauses being in conflict, and the entire contract must be read and understood in order to know whether any other clause conflicts with or overrides the clause we wish to understand. The research question is whether it could be possible to approach contract drafting in a different way, where to understand the meaning of a component (e.g. a clause or expression) it is no longer necessary to read the entire contract. Approaches to modularity include syntactic solutions (e.g. templates, prose objects, common contractual forms)[124] and language solutions (e.g. functional composition,[125] the controlled use of natural language[126]). Another approach is to use non-monotonic reasoning in the semantics, so that the overlap in clauses is directly modelled without the need for modules; for example, Governatori's defeasible deontic logic with violations[127] supports overlapping clauses and provides a superiority relation to manage conflicts.

---

[121] *This example derives from a personal communication with Henning Diedrich.*

[122] *Clack, C. D., & McGonagle, C. (2019). Smart Derivatives Contracts: the ISDA Master Agreement and the automation of payments and deliveries. arXiv preprint arXiv:1904.01461*

[123] *Op. cit. Footnote 58.*

## 5. Summary and Conclusion

For Smart Legal Contracts, where computer technology is used to automate the performance of aspects of legal contracts, a key issue is whether the code is faithful to the contract. This article has exposed some of the complex issues underlying such a seemingly simple question, and in particular it has explored the role played by languages and the problems that arise when translating from high-level languages (such as natural languages) down through the 'language stack' to low-level languages (such as the instruction sets for computers). The problem of validation is not just technical in nature, and a perspective has also been provided on the difficulties that arise when the two specialised fields of law and computer science interact – many of these difficulties also arise in the context of language, such as the large gap in linguistic semantics and pragmatics. This is problematic not just for the automation of legal contracts but also for the current movement towards 'Rules As Code' where it is argued that contracts and legislation should be drafted in both code and natural languages as the same time.[128]

In an attempt to ameliorate some of these issues, a current direction of research within the area of Smart Contracts is the methodology of Computable Contracts, where a single artefact is both the contract (understandable by lawyers who are not programmers) and the code (understandable by computers). An appropriate language for drafting that single artefact must be devised: one that is (i) readily

---

[124] See Section 4.1. Also see Martin, K. (2018). *Deconstructing Contracts: Contract Analytics and Contract Standards. In Data-Driven Law* (pp. 33-34). Auerbach Publications

[125] See Section 4.2, especially the compositional approach: op.cit. Footnote 89

[126] Smith, H.E.: Modularity in contracts: boilerplate and information flow (2006); Michigan Law Review, Volume 104, available at: https://repository.law.umich.edu/cgi/viewcontent.cgi?article=1538&context=mlr

[127] Governatori, G. (2005). Representing business contracts in RuleML. *International Journal of Cooperative Information Systems*, 14(02n03), 181-216.

128 Morris, J. (2021) *Rules as Code: How Technology May change the Language in which Legislation is Written, and What it Might Mean for Lawyers of Tomorrow*, https://s3.amazonaws.com/us.inevent.files.general/6773/68248/1ac865f1698619047027fd22eddbba6e057e990e.pdf See also Marc Lauritsen and Quinten Steenhuis (2019) "Substantive Legal Software Quality: A Gathering Storm?" In ICAIL'19 *Proceedings of the Seventeenth International Conference on Artificial Intelligence and Law*, pp52-62, ACM, and Eyers, J. (2020, Jan 17). Laws should be published in code, says CSIRO. *The Australian Financial Review* generated, understood and used by lawyers, and (ii) expressed using a specification language that may be translated directly to lower layers in the language stack. Various approaches are being pursued, including markup languages, specification languages and domain-specific languages, including controlled natural languages that are also domain-specific languages.

The issue of the drafting lawyer's user experience is not yet resolved. Some Domain Specific Programming Languages (such as CSL[129]) are quite advanced, yet their style is more like programming code than writing legal prose. By contrast, Logical English (see Section 4.3) is a formal specification language that approaches the kind of controlled natural language with which a lawyer might be comfortable, as does Lexon. However, the results of evaluation on large and complex contracts have not yet been published and in their preliminary forms they are not yet as flexible and elegant as uncontrolled English. But these are early days – the preliminary version of Logical English eschews pronouns in order to avoid ambiguity, and although (as observed by Smith[130] and initially by Grice[131]) much of the brevity of natural language comes from the ability to imply context, perhaps the avoidance of ambiguity is more important. Logical English has made initial steps in the implication of context through the controlled use of definite articles, and perhaps a formalisation of more complex implicature in natural language (without re-introducing ambiguity) could be a fruitful direction for future research.

---

[129] *Op. cit. Footnote 94*
[130] *Op. cit. Footnote 126.*
[131] *Grice, H. P. (1975). Logic and conversation. In Speech acts (pp. 41-58). Brill.*